\documentclass{article} 
\usepackage[a4paper,headheight=20pt,top=25mm,bottom=35mm,left=35mm,right=25mm]{geometry}

\usepackage{arxiv}
\usepackage[T1]{fontenc}    
\usepackage{fancyhdr}       
\usepackage{hyphenat}
\usepackage{hyperref}
\usepackage{amsfonts}
\usepackage{amsmath}
\usepackage{amsthm}
\usepackage{graphicx}
\usepackage{subfigure}
\usepackage{enumitem}
\usepackage{array}
\usepackage{booktabs}
\usepackage{natbib}
\usepackage{amssymb}
\usepackage{amsmath}
\usepackage{url}
\usepackage{soul, color}
\usepackage{tikz}

\newcolumntype{C}[1]{>{\centering\let\newline\\\arraybackslash\hspace{0pt}}p{#1}}

\title{Symbolic Explanation of Affinity-Based Reinforcement Learning Agents with Markov Models}

\author{Charl Maree%
\thanks{Second affiliation: Chief Technology Office, Sparebank 1 SR-Bank, Stavanger, Norway.} 
\, and Christian W. Omlin \\ 
Center for Artificial Intelligence Research\\ 
University of Agder\\ 
Grimstad, Norway \\ 
\texttt{\{charl.maree,christian.omlin\}@uia.no} \\ }

\begin{document}
\maketitle

\begin{abstract}
The proliferation of artificial intelligence is increasingly dependent on model understanding. Understanding demands both an interpretation---a human reasoning about a model's behavior---and an explanation---a symbolic representation of the functioning of the model. Notwithstanding the imperative of transparency for safety, trust, and acceptance, the opacity of state-of-the-art reinforcement learning algorithms conceals the rudiments of their learned strategies. We have developed a policy regularization method that asserts the global intrinsic affinities of learned strategies. These affinities provide a means of reasoning about a policy's behavior, thus making it inherently interpretable. We have demonstrated our method in personalized prosperity management where individuals' spending behavior in time dictate their investment strategies, i.e. distinct spending personalities may have dissimilar associations with different investment classes. We now explain our model by reproducing the underlying prototypical policies with discretized Markov models. These global surrogates are symbolic representations of the prototypical policies.
\end{abstract}


\section{Introduction}
The ultimate goal of explainable AI is understanding. It builds trust, improves safety, and improves predictive performance by facilitating precise model improvements \citep{arrieta2020}. For instance, feature saliency analyses can improve feature selection and consequently the predictive performance in stock trading \citep{Carta22}, and rule extraction can enhance trust in an AI system for loan approvals \citet{Sachan20}. Despite considerable advancement in fields such as explainable reinforcement learning (RL) \citep{heuillet2021a, wells21, puiutta20}, the explainability of their underlying models has not yet been fully addressed \citep{Ramon21, Cao21}. 

Reinforcement learning has become omnipotent in finance, for example, multi-agent RL for algorithmic trading \citep{Shavandi22}. Methods such as probabilistic argumentation \citep{riveret19}, structural causal modeling \citep{madumal2019a}, and introspection through interesting elements \citep{sequeira2019a} exemplify the pursuit of post-hoc explainability. We, however, propose an alternative approach: rather than attempting to extract the learned strategy post hoc, ours is an intrinsic method that instills a desirable behavior during training \citep{maree_yourway}. Through regularization of the objective function, our method encourages global action affinities and thus exercises control over what agents learn. We have demonstrated the value of our method in personal prosperity management, where individual spending behaviors dictate investment strategies \citep{maree_can}. We instilled affinities for certain asset classes into the policies of a set of prototypical agents, each associating with a given personality trait. For example, a conscientiousness agent prefers asset classes typically associated with reduced risk. 

Understanding ensues from a model explanation and an interpretation of its behavior. We distinguish between these two concepts: an explanation is a symbolic representation of a model's predictions, while an interpretation is a human reasoning about its behavior. While our agent's policies are inherently interpretable, they lacked a symbolic explanation.  Using discretized Markov models, we now provide that explanation and thus gain insight into previously unanswered questions, such as why do the agents invest according to conventional wisdom: exploiting the benefits of compound growth and reducing risk with increasing customer age. These previously unanswered questions demonstrate the need for both explanations and interpretations: the lack of a symbolic representation of agents' policies inhibited our complete understanding.

Our contributions are therefore: (1) we demonstrate how to instill global action affinities, thus affecting how RL agents learn, which we argue is a useful paradigm shift over the current approach of either post hoc rule extraction or constrained learning, (2) we distinguish between model explainability and interpretability, and in an empirical example demonstrate the difference and the utility of both, and (3) we propose a method of using Markov models to extract symbolic explanations of RL agents' policies. In the next section, we provide an overview of the current state of the art in explainable RL and identify limitations in the field. We then describe our data and empirical methodology, discuss our results, and conclude with insights and future work.

\section{Related Work}
RL agents learn to solve problems by maximizing the total expected reward awarded by the environment in which they act. They are particularly adept at learning in the presence of sparse and delayed rewards \citep{Sutton98}. The environment is a discrete-time process where the current state depends only on the previous state and the action taken by the agent: a Markov decision process (MDP), described by the tuple $(S,A,R,P)$ where $S$ is a set of states, $A$ a set of actions, $R(s,a)$ the reward for taking action $a \in A$ in the state $s \in S$, and $P(s,a) = P(s' \vert s,a)$ the probability that action $a$ in the state $s$ leads to the state $s'$ \citep{bellman}. Deep deterministic policy gradients (DDPG) is a model-free RL algorithm for learning policies in a continuous action space \citep{lillicrap2019}. A DDPG agent consists of four neural networks: an actor $\mu(\theta)$ representing the policy, a critic $Q(\theta)$ representing the state action value function, and for numerical stability, a target actor $\mu'(\theta')$ and a target critic $Q'(\theta')$. During learning, the target network parameters are typically updated slowly given a soft update parameter $\tau \in [0,1]$ with a small value: $\theta'_i = \tau\theta_i + (1-\tau)\theta'_i, \ i \in \{\mu, Q\}$.

Explainable RL has traditionally employed generic methods that explain the underlying models of agents \citep{arrieta2020}. More recently, however, bespoke methods have emerged that consider the state-action space and / or the behavior of the learned policy \citep{puiutta20, heuillet2021a, wells21}. Most, if not all, of these approaches extract explanations after training; they generalize the learned policy through observation or statistical analyses. Few of these extracted explanations match our definition of explainability, and most are more accurately described as interpretations. \emph{State representation learning} connects the state space with information from actions, rewards, or expert knowledge when extracting representations that are useful for reasoning about policies \citep{lesort18}. Under certain restrictions, e.g., linearity, it learns models that either predict states form state-action pairs, or actions from states, thus simplifying the state-action space and improving interpretability. \emph{Introspection} analyzes an agent's experience through statistics such as the frequency of occurrences of states, state-actions, and transitions, the transition probabilities, and estimated rewards compared to the learned state-action value function \citep{sequeira2019a}. It uses interesting elements from this analysis, such as outliers, mean values, etc. to reason about agents' behaviors. \emph{Structural causal modeling} learns causal relationships between states, actions, and rewards by defining action influence graphs that map the action transitions for all possible paths from an initial state to a set of terminal states \citep{madumal2019a}. It defines the causal chain as the one path in the action influence graph that matches the learned policy, and a reward chain as the vector of rewards along this causal chain. Its interpretation of the policy is the comparison between the reward chain and all other possible reward vectors that do not follow the causal chain. \emph{Probabilistic argumentation} uses argumentation graphs---sets of attacking and supporting arguments for each action in a finite action space---to learn interpretations in a RL setting \citep{riveret19}. The state is the intersection of the argumentation graph and the policy to be explained, the actions form a probabilistic distribution across the arguments, and the rewards depend on whether an argument attacks or supports the current action. The learned policy provides probabilistic interpretations of agents' actions in human understandable terms: supporting and attacking arguments for each action. \emph{Reward decomposition} replaces the scalar reward with a vector of more meaningful rewards, where the total reward is the sum of the vector \citep{vanseijen17, juozapaitis2019a, marzari21}. Although evaluating the reward vector for a given action might enable reasoning about that action in meaningful terms, it does not take into account expected future rewards and can be insufficient in environments with delayed or sparse rewards. \emph{Reward redistribution} addresses this problem by redistributing delayed rewards in time; it assigns credit to previous actions, thus reducing the delay of the reward \citep{Dinu2022}. The immediate reward for each time step in a sequence of state-action transitions is equal to the change in the total expected reward. It defines key interpretable events in the policy and, through sequence alignment, redistributes rewards to those events given a set of transition sequences. \emph{Hierarchical RL} divides complex tasks into smaller and simpler tasks that are solved by correspondingly simpler RL agents \citep{beyret2019a, marzari21}. An orchestration agent learns to sequentially combine these prototypical agents to solve complex tasks. If tasks are sufficiently subdivided, the interpretation, or human reasoning about agents' decisions, follows from their simplicity.

The complexity of RL models exacerbates the issue of fidelity and validation of any post hoc explanation. We, instead, encourage agents to adapt their behavior during learning, thus instilling an inherent probabilistic action affinity that is also an interpretation of their behavior \citep{maree_yourway}. Contrary to constrained RL, which avoids certain conditions \citep{Miryoosefi2019, chow2015}, affinity-based learning promotes certain behaviors. This paradigm shift allows the developer to define a desired behavior that an agent must follow during learning, thus instilling a characterization and interpretation during learning; it decouples learned strategies from the reward expectation \citep{aubret19}. Affinity-based RL is not to be confused with preference-based RL that completely eliminates the reward function and instead learns state-action trajectories that maximize the preferences of the expert between pairs of state-action combinations \citep{Wirth17}. Affinity-based RL uses policy regularization that aids---and is never detrimental to---learning convergence by encouraging exploration in environments with complex dynamics or particularly sparse rewards \citep{andres22, Vieillard2020}. It adds a term to the objective function that penalizes any divergence between the current policy and a given prior, for example, Kullback-Leibler (KL) regularization, which uses KL divergence as the distance measure \citep{galashov2018}. Entropy regularization is a specific case of KL-regularization, where the prior is a uniform action distribution that increases the entropy of the policy and thus encourages general exploration of the state-action space\citep{haarnoja2017}. Our method instead encourages exploration of a predefined subset of the state-action space, which describes the desired behavior \citep{maree_yourway}. We define our objective function as follows:
\begin{align} \label{eqn:regularized_obj_func}
    J(\theta) &= \mathbb{E}_{s,a \sim \mathcal{D}} \left[ R(s,a) \right] - \lambda L \\
    L &= \frac{1}{M} \sum_{j=0}^{M} \left[ \mathbb{E}_{a \sim \pi_{\theta}}(a_j) - (a_{j} \vert \pi_{0}(a)) \right]^2 \nonumber
\end{align}
where $\mathcal{D}$ is the replay buffer, $\lambda$ is a hyperparameter that scales the regularization term $L$, $M$ is the number of actions, and $\pi_0$ is a specific prior action distribution that represents the desired behavior. Instilling an interpretable behavior is sufficient for online policy interpretation \citep{Persiani22}.  Unlike KL-regularization, our prior $\pi_0$ is independent of the state and therefore instills a global action affinity in the learned policy. We have demonstrated this in \citet{maree_yourway} where agents navigated a grid towards a destination; they learned to prefer, for example, only right turns and followed optimal paths given their global affinities. In a more elaborate example, we trained a set of prototypical agents with global affinities to invest in certain asset classes \citep{maree_can}. We observed the emergence of interesting investment strategies, such as capitalizing on compound growth and reducing risk with portfolio maturity. Although consistent with conventional wisdom, these strategies were absent from the objective function. To complete our understanding of this behavior, we now provide a symbolic representation---an explanation---of these policies using Markov models.

A hidden Markov model (HMM) models an unobservable Markov process $X$ from its relation to an observable Markov process $Y$; it learns about $X$ by observing $Y$ under the key assumptions that $Y_t$ is solely dependent on $X_t$, and $X_t$ is solely dependent on $X_{t-1}$ (the Markovian property) \citep{Rabiner86}. For a finite hidden state space $X$, there exists a Markov matrix $F$---the sum of the rows add up to one---of state transition probabilities where $F_{ij} = P(X_{n+1} = j \ \vert \ X_n = i)$. Similarly, for a finite observed state space $Y$, there exists a Markov matrix $E$ that describes emission probabilities: $E_{ij} = P(Y_{t} = j \ \vert \ X_t = i)$. We illustrate this process in Figure~\ref{fig:HMM}. Given a series of observed states $\{Y_t\}_{t=0}^T$, the transition and emission probabilities can be estimated using the Baum-Welch algorithm---a special case of the expectation-maximization algorithm \citep{Yang17}.

\usetikzlibrary{matrix}
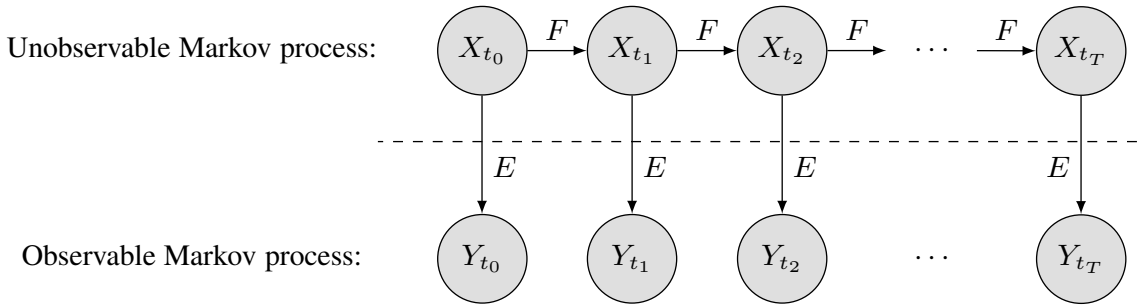
\begin{figure}[!ht]
 \centering
 \footnotesize
 \resizebox{\textwidth}{!}{%
 \begin{tikzpicture}
 \matrix[matrix of math nodes,column sep=2em,row
  sep=4em,cells={nodes={circle,draw,minimum width=3em,inner sep=0pt,fill=gray!25}},
  column 1/.style={nodes={rectangle,draw=none,fill=none}},
  column 5/.style={nodes={rectangle,draw=none,fill=none}},
  ampersand replacement=\&] (m) {
 \text{Unobservable Markov process:} \&
  X_{t_0} \& X_{t_1} \& X_{t_2} \& \cdots \& X_{t_T}\\
 \text{Observable Markov process:} \& 
  Y_{t_0} \& Y_{t_1} \& Y_{t_2} \& \cdots \& Y_{t_T}\\
 };
 \foreach \X in {2,3,4,5}
 {\draw[-latex] (m-1-\X) -- (m-1-\the\numexpr\X+1) node[midway,above]{$F$};
 \ifnum\X=5
 \draw[-latex] (m-1-6) -- (m-2-6) node[pos=0.6,left]{$E$};
 \else
 \draw[-latex] (m-1-\X) -- (m-2-\X) node[pos=0.6,right]{$E$};
 \fi}
 \draw[dashed] ([yshift=1ex]m.east) -- ([yshift=1ex]m.east-|m-1-1.east);
 \end{tikzpicture}}
 \caption{A trellis diagram representing a hidden Markov model with an unobservable Markov process $X$, and observable Markov process $Y$, transition probability matrix $F$, and emission probability matrix $E$.}
 \label{fig:HMM}
\end{figure}

\section{Methodology}
In \citet{maree_personalized}, we defined a set of prototypical agents with intrinsic investment behaviors associated with each of five personality traits: openness, conscientiousness, extraversion, agreeableness, and neuroticism. We used affinity-based RL to learn investment strategies for each of the prototypical agents. Their actions were monthly investment distributions across five different asset classes: savings accounts, property funds, stocks, mortgage curtailment, and luxury items. While stocks, savings, and property investments are self-explanatory, we defined mortgage curtailment as additional payments that reduce the principal debt of a loan, and luxury items such as art, classic cars, fine wines, etc., that might appear in, e.g., the Knight Frank luxury investment index \citep{knight_frank}. We also learned linear combinations of these agents to best match the spending personalities of individual customers which, for the sake of brevity, we do not discuss here. However, to facilitate an understanding of our application, we summarize this paradigm in Figure~\ref{fig:flow_diagram} and refer the reader to a comprehensive account in \citep{maree_personalized}. We now provide an explanation for the prototypical agents' policies using Markov models. 

\tikzstyle{rect} = [rectangle, rounded corners, 
                    text width=2.0cm, 
                    text centered, draw=black, fill=gray!10]
\tikzstyle{circ} = [circle, font=\tiny,
                    text width=0.6cm,
                    text centered, draw=black]
\tikzstyle{frame} = [rectangle, 
                     minimum height=2.3cm,
                     minimum width=2.75cm,
                     draw=black, fill=white!10]
\tikzstyle{frame2} = [rectangle, 
                     minimum height=2.0cm,
                     minimum width=2.3cm,
                     draw=black, fill=white!10]
\tikzstyle{frame3} = [rectangle, 
                     minimum height=2.8cm,
                     minimum width=2.9cm,
                     draw=black, fill=white!10]
\usetikzlibrary{positioning, calc}
\begin{figure}[!ht]
 \centering
 \footnotesize
 \begin{tikzpicture}
  \node (assets) [rect] {Asset prices};
  \node (associ) [rect, below of=assets, yshift=-1.5cm] {Prototypical associations};
  \node (trans) [rect, below of=assets, yshift=-4.9cm] {Transaction history};
  \node (actions) [rect, right of=trans, xshift=9.3cm, yshift=0.1cm] {Generalized investment actions};
  
  \node (rlp) [circ, right of=assets, xshift=1.2cm, yshift=-1.35cm] {RL};
  \node (rnn) [circ, right of=trans, xshift=1.2cm, yshift=0cm] {RNN};
  \node (rlo) [circ, right of=associ, xshift=6.8cm, yshift=1.0cm] {RL};
  \node (rnnf) [circ, below of=rlo, xshift=2.5cm, yshift=-1.7cm] {RNN};
  
  \node (agre) [frame, right of=rlp, xshift=1.63cm, yshift=-0.1cm]{};
  \node (extr) [frame, right of=rlp, xshift=1.58cm, yshift=-0.05cm]{};
  \node (cons) [frame, right of=rlp, xshift=1.53cm, yshift=0cm]{};
  \node (open) [right of=rlp, xshift=1.5cm]{\includegraphics[width=3.2cm]{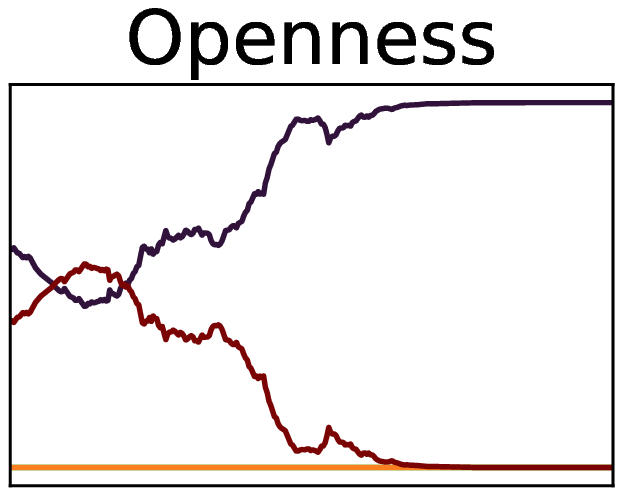}};
  
  \node (feat3) [frame3, right of=rnn, xshift=1.55cm, yshift=-0.1cm]{};
  \node (feat2) [frame3, right of=rnn, xshift=1.5cm, yshift=-0.05cm]{};
  \node (feat1) [frame3, right of=rnn, xshift=1.45cm, yshift=0cm]{};
  \node (feat) [right of=rnn, xshift=1.4cm]{\includegraphics[width=2.8cm]{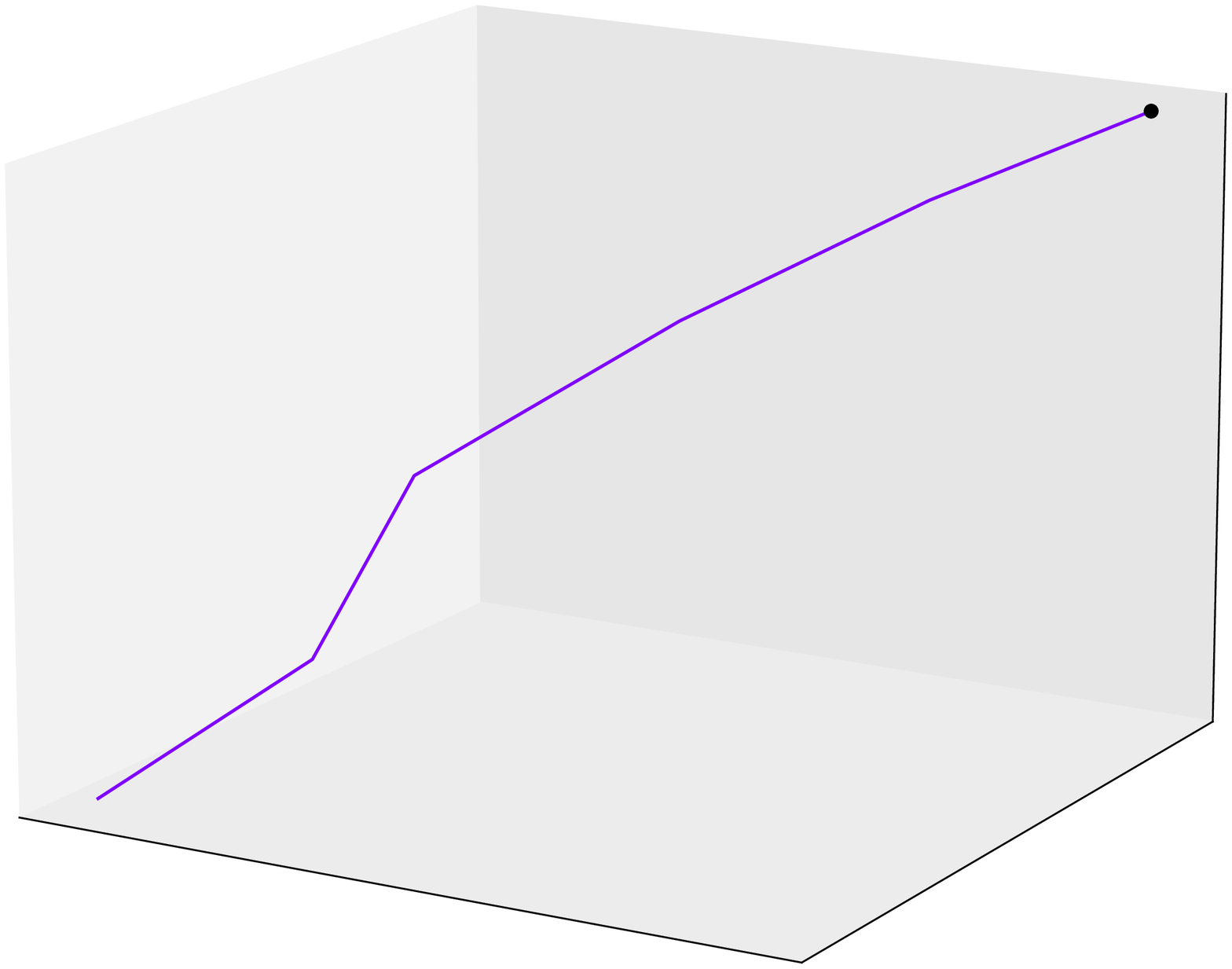}};
  
  \node (orch3) [frame2, right of=rlo, xshift=1.65cm, yshift=-0.1cm]{};
  \node (orch2) [frame2, right of=rlo, xshift=1.6cm, yshift=-0.05cm]{};
  \node (orch1) [frame2, right of=rlo, xshift=1.55cm, yshift=-0.0cm]{};
  \node (orch) [right of=rlo, xshift=1.5cm]{\includegraphics[width=2.7cm]{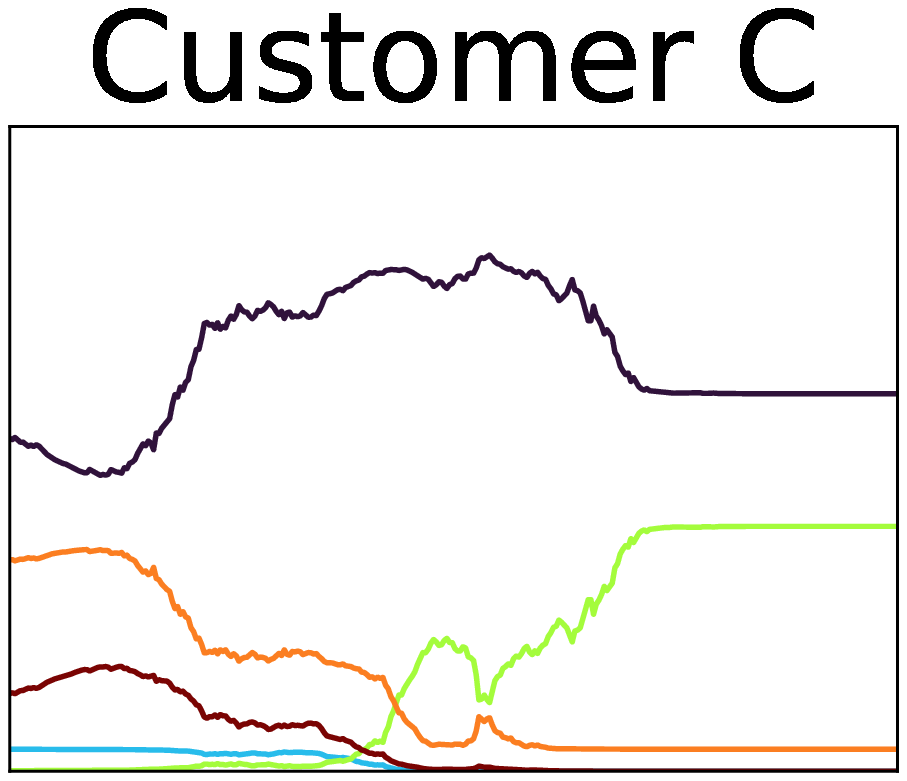}};
  
  \draw[dashed,red,line width=0.5mm] ($(rlp.west) + (-3mm,20mm)$) rectangle ($(open.east) + (1mm,-19mm)$);
  \node[black] at ($(rlp.west) + (21.5mm,16mm)$) {Prototypical investment agents};
  
  \draw[dashed,gray,line width=0.5mm] ($(rlo.west) + (-2mm,20mm)$) rectangle ($(orch.east) + (2mm,-16mm)$);
  \node[black] at ($(rlo.west) + (22mm,16mm)$) {Personal investment strategies};
  
  \draw[dashed,gray,line width=0.5mm] ($(rnn.west) + (-4mm,19mm)$) rectangle ($(feat.east) + (4mm,-18mm)$);
  \node[black] at ($(rnn.west) + (18mm,16mm)$) {Encoded spending behavior};

  \draw[dashed,gray,line width=0.5mm] ($(rnnf.west) + (-15mm,8mm)$) rectangle ($(actions.east) + (5mm,-16mm)$);
  \node[black] at ($(rnnf.west) + (3mm,-28mm)$) {\shortstack{Time-variant generali-\\zation}};
  
  \draw[->] (assets.east) -| ($(assets.east) + (1.5mm,0)$) |- (rlp.west) [black];
  \draw[->] (associ.east) -| ($(assets.east) + (1.5mm,0)$) |- (rlp.west) [black];
  \draw[->] (rlp.east) |- (open.west) [black]; 
  \draw[->] (trans.east) -- (rnn) [black];
  \draw[->] (rnn.east) -- (feat) [black];
  
  \draw[->] ($(open.east) + (0.15cm,0)$) -| ($(rlo.west) - (5mm,0)$) |- (rlo.west) [black];
  \draw[->] ($(feat.east) + (0.4cm,0)$) -| ($(rlo.west) - (5mm,0)$) |- (rlo.west) [black];
  
  \draw[->] ($(feat.east) + (0.4cm,0)$) -| ($(rlo.west) - (5mm,2.8cm)$) |- (rnnf.west) [black];
  \draw[->] (rlo.east) |- (orch.west) [black]; 
  \draw[->] (orch.south) -| (rnnf.north) [black];
  \draw[->] (rnnf.south) -| (actions.north) [black];
  
 \end{tikzpicture}
 \caption{A flow diagram illustrating our system of RL agents that predict personalized investment strategies. There are five prototypical affinity-based RL agents (enclosed in a red dashed rectangle), each associating with one of five personality traits: openness, conscientiousness, extraversion, agreeableness, and neuroticism. These are the agents that we explain using Markov models. Their actions are combined to match the spending behaviors of individual customers, and these combinations are continuously adjusted according to their changing spending behavior using a recurrent neural network (RNN). While these combinations are outside of the scope of this study, we believe it is useful to illustrate how the agents are used in a complete application.}
 \label{fig:flow_diagram}
\end{figure}

\begin{figure}[!ht]
    \centering
    \includegraphics[width=.95\linewidth]{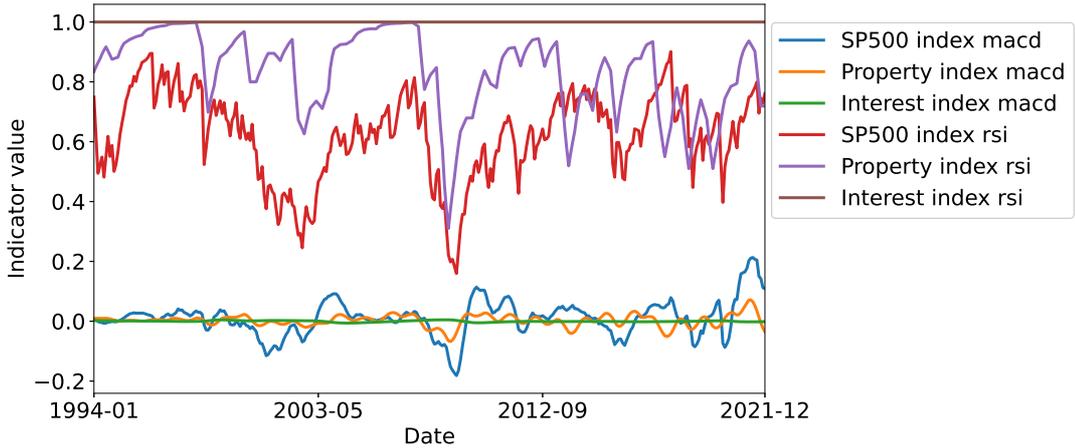}
    \caption{The state data used to train the prototypical agents. We used two common market indicators---the moving average convergence divergence (MACD) and relative strength index (RSI)---to represent market dynamics of the S\&P500 index, Norwegian property index, and Norwegian interest rate index. Our learning time frame was between 1994 and 2022.}
    \label{fig:states}
\end{figure}

To train our agents, we used pricing data for the S\&P500 index, Norwegian property index, and the Norwegian interest rate index between 1994 and 2022. We used two common market indicators---the moving average convergence divergence (MACD) and the relative strength index (RSI) \citep{Chong14}---to capture market dynamics. These indicators are the state space features of the environment in which our agents learned. We show these features in Figure~\ref{fig:states}. There is an additional state variable that indicates the maturity of the portfolio; its value is 0.0 in the first month (January 1994) and linearly increases to 1.0 in the final month (December 2021).

\begin{figure}[!ht]
    \centering
    \includegraphics[width=1.0\linewidth]{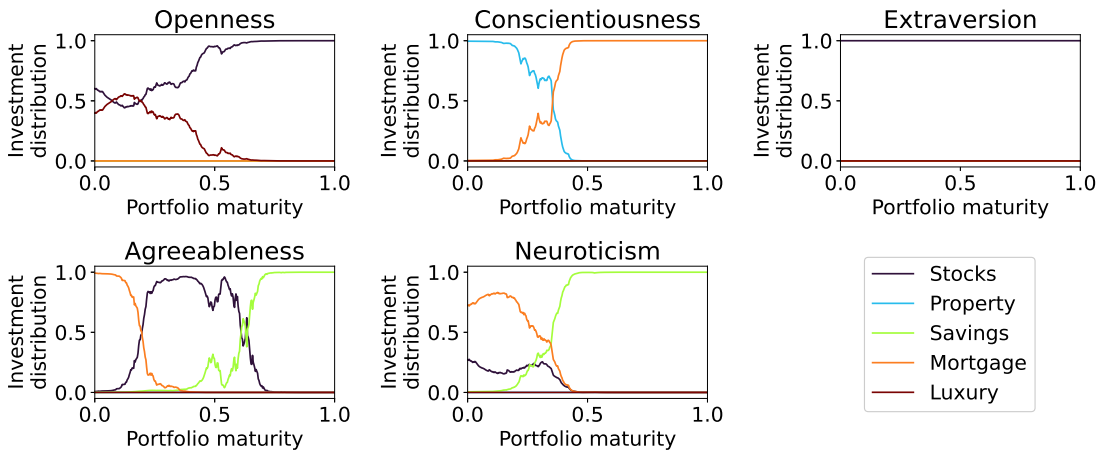}
    \caption{The monthly actions of the five prototypical agents, shown on an x-axis ranging between 0 to 1, representing months between 1994 and 2022. The y-axis represents the monthly investment in each of the asset classes. Note that the actions strictly represent the purchase of assets, i.e. the extraversion agent, for instance, consistently invests 100\% of available monthly funds into stocks, thus consistently increasing the portfolio holding of stocks; assets are never sold. Though the agents optimised a common reward function---monthly returns---, their distinct strategies were instilled through affinity-based learning. }
    \label{fig:proto_actions}
\end{figure}

We show the resulting policies for the five prototypical agents in Figure~\ref{fig:proto_actions}. The agents optimized a common reward function, i.e., monthly returns; they maximized the portfolio value. Though they shared a common reward function, the agents learned unique investment strategies: the conscientiousness agent, for instance, prefers low-risk investment in property followed by resolute mortgage curtailment, while the openness agent prefers investments that might incite their curiosity, such as luxury items and stocks. 

To train Markov models that match the predictions of the five prototypical agents, we discretized the states and actions of the agents. We assigned three bins to the RSI indicator based on the knowledge that values between 0 and 0.3 indicate oversold conditions, values between 0.7 and 1 indicate overbought conditions, and values between 0.3 and 0.7 are inconclusive \citep{Chong14}. We similarly assigned two bins for the MACD indicator based on the knowledge that positive values represent a buy signal, while negative values represent a sell signal \citep{Chong14}. We divided the maturity state feature into 28 bins: one for each year of the investment period. We finally assigned 5 equally sized bins for the agents actions, between 0 and 1. This resulted in 168 potential states, of which only 102 states ever occurred. It is reasonable that not all possible states occurred, since MACD and RSI are related; it is not unexpected that whenever RSI indicates oversold conditions, MACD could suggest a buy signal \citep{Chong14}. We then estimated the transition probabilities in the Markov matrices $F_i$ and the emission probabilities $E_i$, where $i \in [1,5]$, for the five Markov models by observing the state transitions and the corresponding actions for each of the prototypical agents. Using the initial state and the five Markov models defined by $F_i$ and $E_i$, we can reproduce the policies of the five prototypical agents with high fidelity.

\section{Results}
\begin{figure}[!hpt]
    \centering
    \includegraphics[width=.70\linewidth]{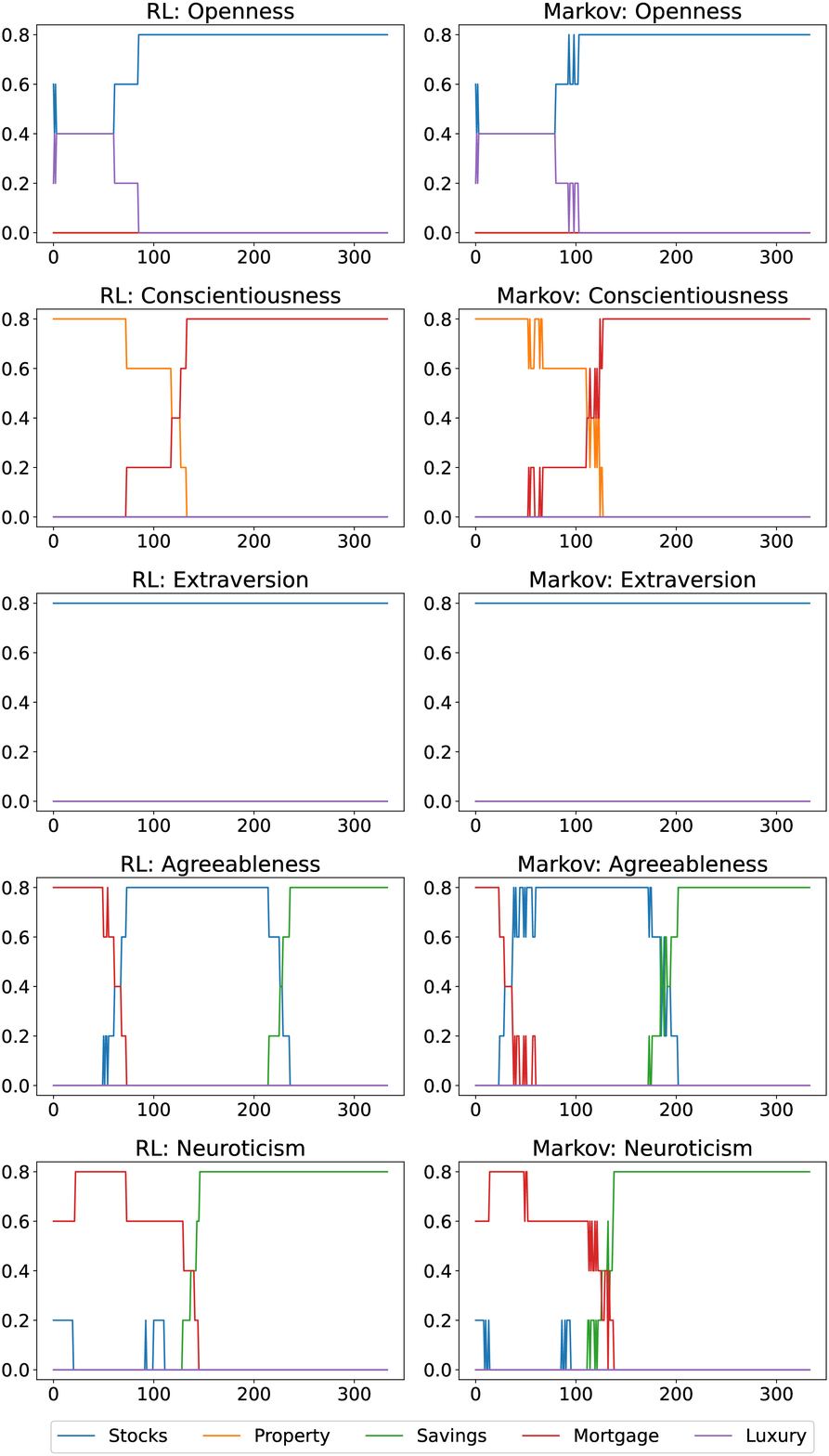}
    \caption{A visual comparison between the discretized predictions of five RL agents (on the left) and the five corresponding Markov models (on the right). The single input to the Markov models is the initial state, from which they predict the transition to the next state and the corresponding action by the agent. The Markov models clearly predict the actions with high fidelity.}
    \label{fig:markov_performance}
\end{figure}

We trained five distinct Markov models as global surrogates to reproduce the predictions of five affinity-based RL agents. We show the discretized actions of the agents and the corresponding predictions of the Markov models in Figure~\ref{fig:markov_performance}. 
Using only the initial state as input, the Markov models predict the agents' actions with high fidelity, with some uncertainly when action values change due to the probabilistic nature of Markov models. 

\begin{figure}[!ht]
    \centering
    \includegraphics[width=.99\linewidth]{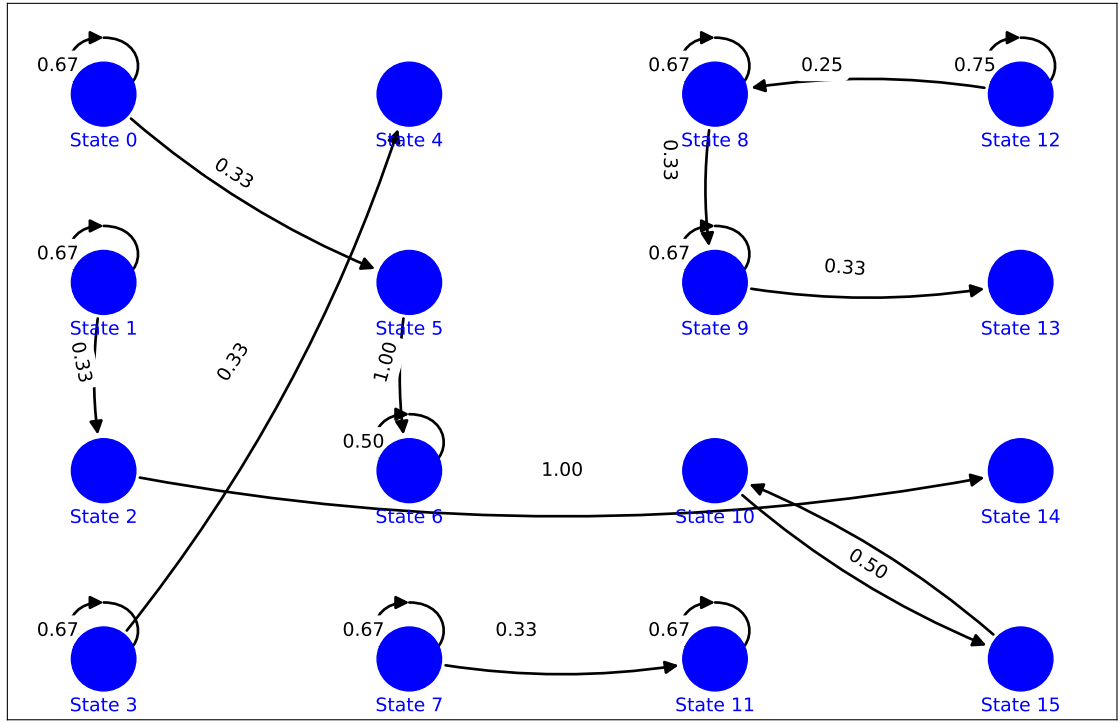}
    \caption{A non-exhaustive illustration of the trained Markov model showing state transitions for a subset of states. States are shown as blue circles, and state transitions and their probabilities are shown in black. We show the first 16 states, as visualizing all 102 states is not feasible. Each state represents a set of features with discretized values for MACD and RSI indicators of the property and stock indices, respectively, as well as the maturity of the portfolio. Note that not all state transitions are shown, since the origin or destination state might not be included in this subset.}
    \label{fig:state_transitions}
\end{figure}

Figure~\ref{fig:state_transitions} shows the state transitions for a non-exhaustive subset of states: the first 16 states visited including the initial state. 
We observe that not all states are visited, which is expected since the market indicators MACD and RSI are not entirely independent, nor are the stock and property markets in general. For example, during macroeconomic downturns we often observe a decline in both these markets: refer to Figure~\ref{fig:states} and observe, for example, the decline in both the property and S\&P500 indices during the 2008 recession. Property and stock markets can also demonstrate an inverse correlation: in Figure~\ref{fig:states} the RSI curves for property and stocks can have reversed slopes, while the MACD curve can exist on opposite sides of zero. By perturbing the sizes and number of bins, we observed that portfolio maturity holds the most salient information. This is an important observation; it suggests that the values of the market indicators have a lesser influence on investment strategies compared to the maturity of the portfolio. This is in line with conventional wisdom that long-term investment should not be overly concerned with short-term market volatility; property and stock indices have typically followed an upward trend in the long run. The reduced dependence on market conditions increases confidence in model robustness when trading on unseen data: the unseen market conditions are less important than investor age; the basic principle that younger investors can afford increased risk in return for higher reward, and mature investors should seek to reduce portfolio risk, is common across a wide range of market conditions. 

\section{Conclusions}
Understanding deep AI models requires an interpretation of their behavior and a symbolic representation, or explanation, of their functioning. These two elements facilitate reasoning about a model and, thus, enhance trust in its decisions. We have proposed a novel affinity-based approach to interpretable reinforcement learning; it encourages exploration of a predefined subset of the state-action space. This prior action distribution describes the agent's desired behavior and is the interpretation of its policy. However, our solution lacked a symbolic explanation, resulting in unanswered questions about why they make certain decisions. A concrete example is why a set of agents, that learned to invest according to the preferences of prototypical personality traits, invest in more risky assets for younger investors and reduce risk with investor age. We now provide a symbolic representation of the agents' policies, using Markov models, that answer such questions. Our Markov models recreate, with high fidelity, the discretized investment strategies of five prototypical investment agents using only the initial state. By perturbing the bin sizes of the discretized state features, we are able to determine the most salient feature: portfolio maturity. The fact that market conditions play a diminutive role in model prediction is significant: it enhances trust in out-of-sample predictions and suggests that investment timing is more important than market conditions. The agents make use of compounding growth by investing in higher reward---but more risky---assets early on, and fulfill their prescribed action distributions towards the end of the investment period; they learned how to maximize rewards. This use case demonstrates the need for both interpretations and explanations to fully comprehend the functioning and characterization of deep RL systems. The Markov model is a valuable tool for extracting a symbolic representation of an otherwise opaque RL model, and affinity-based RL is a unique approach to control what RL agents learn and thus interpret their behavior. It is a paradigm shift from current approaches that either encourage general exploration for the purpose of improved convergence or constrain the state space to prevent the policy from visiting undesirable states. It is compelling to apply affinity-based RL to virtuous agents, personalized learning and teaching, chronic disease treatment, climate change, wind farm operations, etc.

\section*{Declaration of competing interest}
The authors declare that they have no competing interests.
\section*{Funding}
This study was partially funded by a grant from the Norwegian Research Council, project number 311465.

\bibliographystyle{unsrtnat}
\bibliography{references}

\end{document}